\documentclass[twoside]{article}
\usepackage{aistats2020}
\usepackage{float}

\usepackage[separate-uncertainty]{siunitx}
\usepackage{subcaption}
\usepackage{multirow}
\usepackage{booktabs}
\usepackage{mathtools,amssymb,amsthm,amsfonts,amssymb,amscd,mathtools,bm}

\usepackage{float}
\usepackage{xcolor}

\DeclarePairedDelimiter\parens{\lparen}{\rparen}

\DeclarePairedDelimiterX{\infdivx}[2]{\lparen}{\rparen}{#1\;\delimsize\|\;#2}

\newcommand{\Normaldist}[1]{\mathcal{N}\parens*{#1}}
\newcommand{\E}[2][]{
   \ifthenelse{ \equal{#1}{} }
      {\ensuremath{\left\langle#2\right\rangle}}
      {\ensuremath{\left\langle#2\right\rangle_{#1}}}
}
\newcommand{\cov}[2][]{
    \ifthenelse{ \equal{#1}{} }
        {\operatorname{cov}\parens*{#2}}
        {\operatorname{cov}_{#1}\parens*{#2}}
}

\PassOptionsToPackage{
    style=authoryear-comp,
    hyperref=false,
    backend=biber,
    maxcitenames=2,
    uniquename=false,
    url=false,
    doi=false,
    isbn=false,
    eprint=false,
}{biblatex}
\usepackage{biblatex}

\begin{document}
\twocolumn[
    \aistatstitle{Learning GPLVM with arbitrary kernels\\using the unscented transformation}
    \aistatsauthor{
        Daniel Augusto de Souza$^1$ \And
        Diego Mesquita$^2$ \And
        César Lincoln Mattos$^1$ \And
        João Paulo Gomes$^1$
    }
    \vspace{0.75cm}
]
\begin{abstract}
Gaussian Process Latent Variable Model (GPLVM) is a flexible framework to handle uncertain inputs in Gaussian Processes (GPs) and incorporate GPs as components of larger graphical models. Nonetheless, the standard GPLVM variational inference approach is tractable only for a narrow family of kernel functions. The most popular implementations of GPLVM circumvent this limitation using quadrature methods, which may become a computational bottleneck even for relatively low dimensions. For instance, the widely employed Gauss-Hermite quadrature has exponential complexity on the number of dimensions.  In this work, we propose using the unscented transformation instead. Overall, this method presents comparable, if not better, performance than off-the-shelf solutions to GPLVM and its computational complexity scales only linearly on dimension. In contrast to Monte Carlo methods, our approach is deterministic and works well with quasi-Newton methods, such as the Broyden-Fletcher-Goldfarb-Shanno (BFGS) algorithm. We illustrate the applicability of our method with experiments on dimensionality reduction and multistep-ahead prediction with uncertainty propagation.
\end{abstract}

\section{INTRODUCTION}

Gaussian process (GP) models have been widely adopted in the machine learning community as a Bayesian approach to nonparametric kernel-based learning due to their simplicity and fully probabilistic predictions \parencite{rasmussen2006}. Thanks to its flexibility, many authors have applied the GP framework in contexts such as dynamical modeling \parencite{mattos2016_iclr,eleftheriadis2017identification}, autoencoders \parencite{eleftheriadis2016variational,casale2018gaussian}, and hierarchical modeling \parencite{salimbeni2017doubly,havasi2018inference}.

The works above have a significant building block in common: the GP Latent Variable Model (GPLVM), which was proposed by \textcite{lawrence2004} to handle learning scenarios with uncertain inputs. The GPLVM was extended with a Bayesian training approach (Bayesian GPLVM) by \textcite{titsias2010} and later by \textcite{damianou2013} in a multilayer setting (Deep GPs).

The variational approach presented by \textcite{titsias2010} for the Bayesian GPLVM presents tractable calculations only for a few choices of kernel function, such as the radial basis function (RBF) kernel. However, it is known that the RBF kernel presents limited extrapolation capability \parencite{mackay1998introduction}. Some authors have tried to address that issue. The work by \textcite{duvenaud2013structure,lloyd2014automatic} pursues a compositional approach to build more expressive kernels from simpler ones. \textcite{wilson2013gaussian} propose the spectral mixture kernel family, capable of automatic pattern discovery and extrapolating beyond the training data. \textcite{wilson2016deep,wilson2016stochastic,al2017learning} propose using deep neural networks to learn kernel functions directly from the available data. Although those proposals achieve more flexible models than those with the RBF kernel, they turn some Bayesian GPLVM expressions intractable.

Some works, such as \textcite{eleftheriadis2017identification,salimbeni2017doubly}, handle non-RBF kernels with uncertain inputs using the so-called ``reparametrization trick'' \parencite{KingmaWelling2014, RezendeEtAl2014} in the doubly stochastic variational inference framework, introduced by \textcite{titsias2014doubly}.
This approach results in a flexible inference methodology, but, by straying from the deterministic variational methods, it does not support inference using popular quasi-Newton methods, like the Broyden–Fletcher–Goldfarb–Shanno (BFGS) algorithm.

In the present paper, we aim to handle the propagation of uncertainty in the GPLVM while maintaining the non-stochastic framework presented by \textcite{titsias2010}. We tackle the intractabilities of uncertain inputs and non-RBF kernels by employing the unscented transformation (UT), a deterministic technique to approximate nonlinear mappings of a probability distribution \parencite{julier2004unscented,menegaz2015systematization}. The UT projects a finite number of \textit{sigma points} through a nonlinear function and uses their computed statistics to estimate the transformed mean and covariance, resulting in a more scalable method than, for instance, the Gauss-Hermite (GH) quadrature.

We use the UT to handle the intractabilities of the Bayesian GPLVM and propose using this approximation in the integrals that arise by convolving kernel functions and a Gaussian density in the variational framework by \textcite{titsias2010}. Our methodology enables the use of any kernel, including ones obtained via auxiliary parametric models in a kernel learning setup, while maintaining fast deterministic inference. We evaluate this approach in GPLVM's original task of dimensionality reduction and in the task of uncertainty propagation during a free simulation (multistep-ahead prediction) of dynamical models. Our experimental results show that, even for a moderate latent space size, the commonly used GH quadrature is only feasible when the user picks a very low number of evaluation points. Moreover, in such scenarios, the UT still presents excellent results.


In summary, our main contributions are: (\textit{i}) an extension to the Bayesian GPLVM using the UT to handle intractable integrals deterministically and enable the use of any kernel; (\textit{ii}) a set of experiments comparing the proposed approach and alternative approximations using Gauss-Hermite quadrature and Monte Carlo sampling in tasks involving dimensionality reduction and dynamical free simulation.

The remainder of the paper is organized as follows. In Section \ref{SEC_THEORY} we present the theoretical background by summarizing the GPLVM framework and the UT approximation. In Section \ref{SEC_PROPOSE} we detail our proposal to apply the UT within the Bayesian GPLVM setting. In Section \ref{SEC_EXP} we present and discuss the obtained empirical results. Finally, in Section \ref{SEC_REL} we review related works related to GPs and UT and we conclude the paper in Section \ref{SEC_CONC} with ideas for further work.
\newpage

\section{THEORETICAL BACKGROUND}
\label{SEC_THEORY}

In this section, we summarize the GP and the Bayesian GPLVM models, as well as the UT.

\subsection{The Gaussian Process Framework}

Let $N$ inputs $\bm{x}_i \in \mathbb{R}^{D_x}$, organized in a design matrix $\bm{X} \in \mathbb{R}^{N \times D_x}$ be mapped via $f : \mathbb{R}^{D_x} \rightarrow \mathbb{R}^{D_y}$ to $N$ correspondent outputs $\bm{f}_i \in \mathbb{R}^{D_y}$, organized in the matrix $\bm{F} \in \mathbb{R}^{N \times D_y}$. We observe $\bm{Y} \in \mathbb{R}^{N \times D_y}$, a noisy version of $\bm{F}$. Considering an observation noise $\bm{\epsilon} \sim \mathcal{N}(\bm{0}, \sigma^2\bm{I})$, we have $\bm{f}_{:d} = [f(\bm{x}_1) \ldots f(\bm{x}_N)]^\top$ and $\bm{y}_{:d} = \bm{f}_{:d} + \bm{\epsilon}$, where $\bm{y}_{:d} \in \mathbb{R}^N$ is comprised of the $d$-th component of each observed sample, i.e., the $d$-th column of the matrix $\bm{Y}$. If we choose independent multivariate zero mean Gaussian priors for each dimension of $\bm{F}$, we get \parencite{rasmussen2006}:
\begin{equation*}
p(\bm{Y}|\bm{X}) = \prod_{d=1}^{D_y} \mathcal{N}( \bm{y}_{:d} | \bm{0}, \bm{K}_f + \sigma^2 \bm{I}),
\end{equation*}
where we were able to analytically integrate out the non-observed (\textit{latent}) variables $\bm{f}_{:d}|_{d=1}^{D_y}$. The elements of the covariance matrix $\bm{K}_f \in \mathbb{R}^{N \times N}$ are calculated by $[\bm{K}_f]_{ij} = k(\bm{x}_i, \bm{x}_j), \forall i,j \in \{1,\cdots,N\}$, where $k(\cdot, \cdot)$ is the so-called covariance (or \textit{kernel}) function.

\subsection{The Bayesian GPLVM}
\label{section_BGPLVM}

The Gaussian Process Latent Variable Model (GPLVM), proposed by \textcite{lawrence2004}, extends the GP framework for scenarios where we do not observe the inputs $\bm{X}$, which generated the response variables $\bm{Y}$ via the modeled function. 
The GPLVM was originally proposed in the context of nonlinear dimensionality reduction\footnote{The GPLVM is a nonlinear extension of the probabilistic Principal Component Analysis \parencite{lawrence2004}.}, which can be done choosing $D_x < D_y$. However, the approach has proved to be flexible enough to be used in several other scenarios. For instance, in supervised tasks, the matrix $\bm{X}$ can be seen as a set of observed but uncertain inputs \parencite{damianou2016variational}.

The Bayesian GPLVM, proposed by \textcite{titsias2010}, considers a variational approach \parencite{jordan1999} to approximately integrate the latent variables $\bm{X}$. Inspired by Titsias' variational sparse GP framework \parencite{titsias2009}, the Bayesian GPLVM avoids overfitting by considering the uncertainty of the latent space and enables the determination of $D_x$ by using a kernel function with ARD (\textit{automatic relevance determination}) hyperparameters.



Following \textcite{titsias2010}, we start by including $M$ inducing points $\bm{u}_{:d} \in \mathbb{R}^M$ associated to each output dimension and evaluated in $M$ pseudo-inputs $\bm{z}_j|_{j=1}^M \in \mathbb{R}^{D_x}$, where $p(\bm{u}_{:d}) = \mathcal{N}(\bm{u}_{:d} | \bm{0}, \bm{K}_z)$ and $\bm{K}_z \in \mathbb{R}^{M \times M}$ is the kernel matrix computed from the pseudo-inputs. The joint distribution of all the variables in the GPLVM $p(\bm{Y}, \bm{X}, \bm{F}, \bm{U})$ is now given by (with omitted dependence on $\bm{z}_j$):%
\begin{equation*}
p(\bm{X})\prod_{d=1}^{D_y} p(\bm{y}_{:d} | \bm{f}_{:d}) p(\bm{f}_{:d} | \bm{u}_{:d}, \bm{X}) p(\bm{u}_{:d}).
\end{equation*}
Applying Jensen's inequality to the above expression gives a lower bound to the marginal log-likelihood $\log p(\bm{Y})$:
\begin{align*}
\log p(\bm{Y}) &= \log \int p(\bm{Y}, \bm{X}, \bm{F}, \bm{U}) \mathrm{d}\bm{X} \mathrm{d}\bm{F} \mathrm{d}\bm{U}\\&\geq \int Q \log \left[ \frac{p(\bm{Y}, \bm{X}, \bm{F}, \bm{U})}{Q} \right] \mathrm{d}\bm{X} \mathrm{d}\bm{F} \mathrm{d}\bm{U},
\end{align*}
where $Q$ is the variational distribution, chosen to be given by the form $Q = q(\bm{X}) q(\bm{U}) p(\bm{F} | \bm{U}, \bm{X})$, where $p(\bm{F} | \bm{U}, \bm{X})$ is an analytical conditional distribution of Gaussians and the variational distributions $q(\bm{X}) = \prod_i^N q(\bm{x}_i)$ and $q(\bm{U}) = \prod_d^{D_y} q(\bm{u}_{:d})$ respectively approximate the posteriors of the variables $\bm{X}$ and $\bm{U}$ by products of multivariate Gaussians.

The final analytical bound derived in \textcite{titsias2010}, which may be directly used to perform model selection, depends on the three terms, named $\Psi$-statistics. Those terms, which will be presented in the next section, represent convolutions of the kernel function with the variational distribution $q(\bm{X})$ and are tractable only for a few kernel functions, such as the RBF, the linear kernels, and their mixtures.

\subsection{The Unscented Transformation}
\label{sec_ut}

The unscented transformation (UT) is a method for estimating the first two moments of a transformed random variable under an arbitrary function. First proposed by \textcite{uhlmann1995dynamic} for non-linear Kalman filters, the transformation itself is decoupled from the proposed Unscented Kalman Filter.

In the UT, the mean and covariance of the transformed random variable are approximated with a weighted average of transformed \textit{sigma points} $\bm{S}$, derived from the first two moments of the original input.

Let $p(\bm{x}) = \Normaldist{\bm{\mu},\bm{\Sigma}}$, where $\bm{x}\in \mathbb{R}^D$, be the input of an arbitrary transformation $\bm{f}\colon \mathbb{R}^D \rightarrow \mathbb{R}^Q$. Given uniform weights for the sigma points, the output moments are computed by:
\begin{align}
    \label{eq_ut_mean}
    \E[p(\bm{x})]{\bm{f}(\bm{x})} &\approx \frac{\textstyle\sum_i^{2D} \bm{f}(\bm{s}_i)}{2D} = \bm{\tilde{\mu}}_\mathrm{UT},\\
    \nonumber
    \mathbf{Cov}(\bm{f}(\bm{x})) &\approx \frac{\textstyle\sum_i^{2D} (\bm{f}(\bm{s}_i)-\bm{\tilde{\mu}}_\mathrm{UT})(\bm{f}(\bm{s}_i)-\bm{\tilde{\mu}}_\mathrm{UT})^\intercal}{2D}.
\end{align}

There are several strategies to select sigma points\footnote{e.g. \parencite{menegaz2015systematization}}, however, we follow the original scheme by \textcite{uhlmann1995dynamic}, with uniform weights and sigma points chosen from the columns of the squared root of $D\bm{\Sigma}$, an efficient way to generate a symmetric distribution of sigma points.

This scheme is defined as follow. Let $\mathbf{Chol}(\bm{\Sigma})$ be the Cholesky decomposition of the matrix $\bm{\Sigma}$. Then, the sigma points $\bm{S}$ are defined as:
\begin{align*}
    \bm{s}_i &= \bm{\mu} + [\mathbf{Chol}(D\bm{\Sigma})]_{:i}\\
    \bm{s}_{i+D} &= \bm{\mu} - [\mathbf{Chol}(D\bm{\Sigma})]_{:i}, \quad \forall i \in [0,D),
\end{align*}
where $[\mathbf{Chol}(D\bm{\Sigma})]_{:i}$ denotes the $i$-th column of the matrix $\mathbf{Chol}(D\bm{\Sigma})$.

\section{PROPOSED METHODOLOGY}
\label{SEC_PROPOSE}


This section details our proposal, discusses its advantages and limitations, and presents an initial empirical validation.

\subsection{Learning Bayesian GPLVMs using UT}

As mentioned in Section \ref{section_BGPLVM}, the computation of the $\Psi$-statistics is the only part that prevents the application of the Bayesian GPLVM with arbitrary kernels. We propose to tackle such computations in intractable cases using the mean approximated by the UT (see Eq. \eqref{eq_ut_mean}), as follows:

\begin{align}
\label{EQ:PSI_0}
&\psi_0 = \sum_i^N\E[q(\bm{x}_i)]{k(\bm{x}_i,\bm{x}_i)} \approx \frac{1}{2D}\sum_k^{2D} k(\bm{s}_k, \bm{s}_k),\\
\label{EQ:PSI_1}
&[\bm{\Psi}_1]_{ij} =
\E[q(\bm{x}_i)]{k(\bm{x}_i,\bm{z}_j)} \approx \frac{1}{2D}\sum_k^{2D} k(\bm{s}_k, \bm{z}_j), \\
\label{EQ:PSI_2}
&\begin{aligned}
    [\bm{\Psi}_2]_{jm} &= \sum_i^N \E[q(\bm{x}_i)]{k(\bm{x}_i,\bm{z}_j)k(\bm{x}_i,\bm{z}_m)} \\
    &\approx \frac{1}{2D}\sum_i^N\sum_k^{2D} k(\bm{s}^{(i)}_k, \bm{z}_j)k(\bm{s}^{(i)}_k, \bm{z}_m)
\end{aligned},
\end{align}
where $\psi_0\in\mathbb{R}$, $\bm{\Psi}_1 \in \mathbb{R}^{n \times m}$, and $   \bm{\Psi}_2 \in \mathbb{R}^{m \times m}$ are the $\Psi$-statistics and $\bm{s}^{(i)}_k$ indicates the $k$-th sigma point related to $q(\bm{x}_i)$.

\subsection{Advantages and limitations}

Besides enabling the use of non-analytical kernels in the Bayesian GPLVM, the choice of using UT-based approximations in place of, for instance, the Gauss-Hermite (GH) quadrature, brings great computational benefits, due to the number of points that are evaluated to compute the Gaussian integral.
Given a $D$-dimensional random variable, the UT requires just a linear number of $2D$ evaluations, while the GH quadrature requires $H^D$ evaluations, where $H$ is a user-chosen order parameter. Even for $H=2$ and moderate dimensionality values, e.g. $D=20$, the GH approach would require at least $2^{20}$ evaluations per approximation, which is infeasible. 

Since an exponentially lower number of function evaluations is required, the UT presents a practical alternative to the GH quadrature. Furthermore, since the sigma points are obtained in a fully deterministic manner, it enables quasi-Newton optimization methods, unlike Monte Carlo integration. Nevertheless, if a large quantity of evaluations is allowed for either GH quadrature or Monte Carlo (MC) integration, in exchange for additional computational effort, it is expected that the lower number of sigma points of the UT would result in a coarser approximation.

Regarding the approximation quality, \textcite{menegaz2015systematization} proved that our choice for sigma points (detailed in Section \ref{sec_ut}) enables computing the projected mean correctly up to the third-order Taylor series expansion of the transformation function if $\bm{x}$ is Gaussian distributed. Note that for the GH quadrature the approximation is guaranteed up to the $(2H-1)$th order of the function. So, for the $H=2$ case, it is expected that both approximations will have about the same quality.



\subsection{Preliminary validation}

In the Bayesian GPLVM, the amount of sampled points is relevant, since the approximations are computed at each step of the variational lower bound optimization. Thus, the number of times we evaluate the $\Psi$-statistics gives a raw estimate of the chosen approximation computational budget.

To verify how performance evolves with dimensionality when using UT in the context of the Bayesian GPLVM, we computed $\bm{\Psi}_1$ -- see Eq. \eqref{EQ:PSI_1} -- considering a RBF kernel on random data ($N=40, M=20$) of varying dimension. We compare the UT result with the GH quadrature and MC integration.

\begin{figure}
    \centering
    \includegraphics[width=\linewidth]{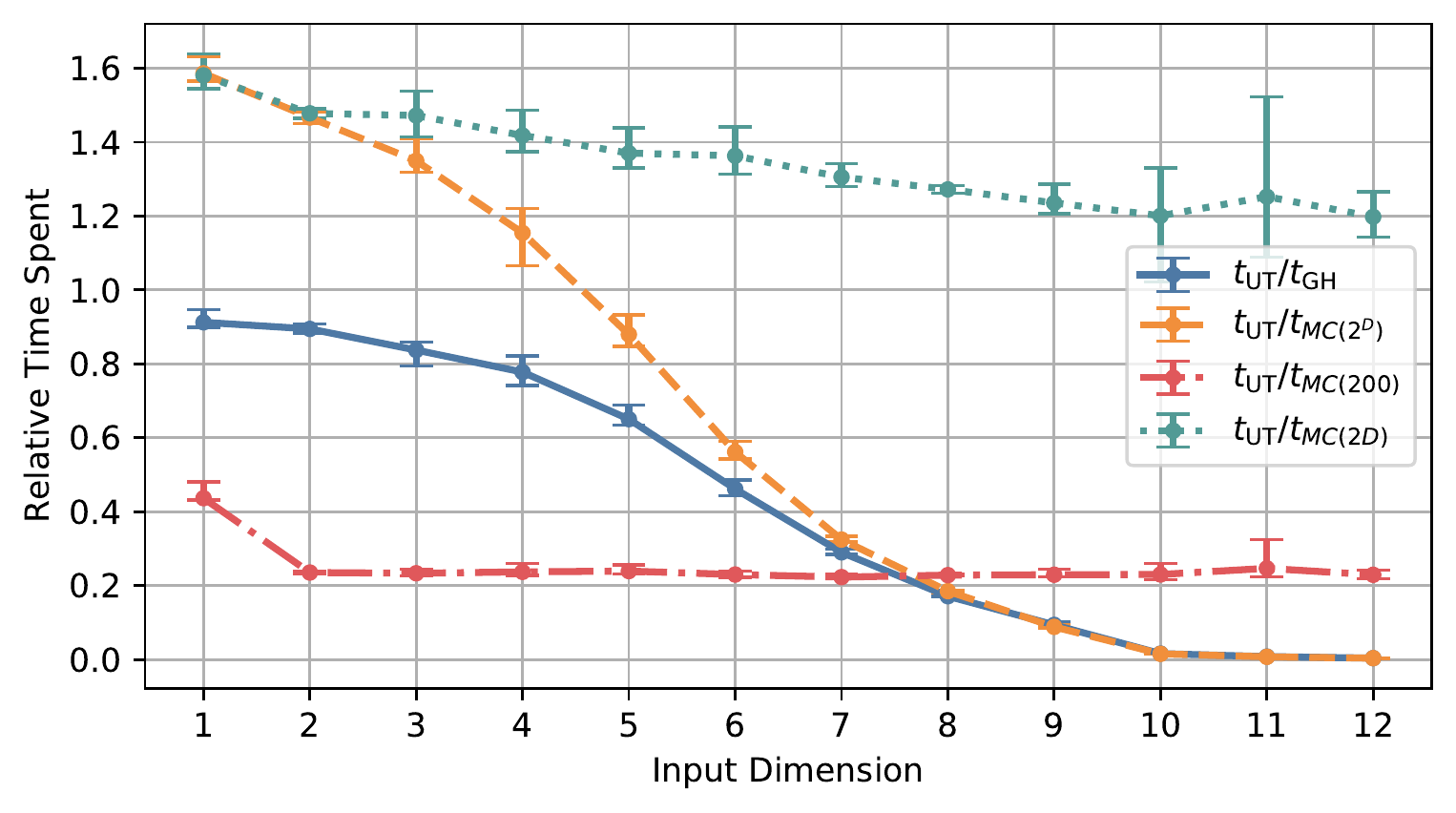}
    \caption{Relative computational effort of the UT with other methods when computing $\Psi$-statistics. In practice, even in low dimensions, the overhead of computing the Hermite roots and weights can make GH slower than UT. As expected, the exponential nature of GH makes it infeasible at dimensions beyond 10.}
    \label{fig:ut_time}
\end{figure}

Figure \ref{fig:ut_time} shows a comparison of the relative time spend between the UT and four competing quadrature methods: GH and MC with $2D$, $2^D$, and $200$ samples. As expected from the theoretical complexity, GH's exponential nature makes it infeasible at dimensions beyond 10. Not only that, but the complexity of GH brings an additional overhead that is apparent even in small dimensions. On the other hand, MC integration's simplicity brings its runtime to be faster than UT on small dimensions, but, as it will become apparent in the next section, at this regime, the UT can achieve better results even when compared to the MC with 200 samples.

\section{EXPERIMENTS}
\label{SEC_EXP}

This section intends to evaluate the UT against other approximations methods, showing its practicability in quality and speed on tasks requiring solving Psi-statistics during model training and model prediction. We considered two standard tasks for the GPLVM that fit this criterion: dimensionality reduction and free simulation of dynamical models with uncertainty propagation.

We compared the proposed UT approach with the GH quadrature and the reparametrization trick based MC sampling for computing the $\Psi$-statistics of the Bayesian GPLVM. In the tractable cases, we also considered the analytical expressions. All experiments were implemented in Python using the GPflow framework \parencite{matthews2017gpflow}. The code can be found at
\url{https://github.com/spectraldani/UnscentedGPLVM}.

For the GH experiments, to maintain a reasonable computational cost, we used $2^D$ points, where $D$ is the input dimension. For the MC approximations, we used three different numbers of samples: the same number used by UT, the same number used by GH, and a fixed quantity of 200 samples. Each MC experiment was run ten times, with averages and standard deviations reported. The MC approximation is similar to the one in the doubly stochastic variational framework \parencite{titsias2014doubly}, but without mini-batch updates.

The kernel hyperparameters, likelihood noise, and variational parameters are all joint\-ly optimized using the second-order optimization method L-BFGS-B \parencite{Byrd1995}. However, it is not feasible to use L-BFGS-B for the models with MC sampling, so, these models were optimized using Adam \parencite{KingmaWelling2014} with a learning rate of 0.01.

\subsection{Dimensionality Reduction}

The dimensionality reduction task is especially suitable for the UT-based approach since the dimension of the integrand in the $\Psi$-statistics are usually small for data visualization purposes.

We used two datasets, which were referred in \textcite{lawrence2004} and \textcite{titsias2010}, the Oil flow dataset, and the USPS digit dataset. In both cases, we compared the analytic Bayesian GPLVM model with the RBF kernel against a kernel with non-analytic $\Psi$-statistics. The following kernels were considered: RBF, Matérn 3/2 and a Multilayer Perceptron (MLP) composed on a RBF kernel, similar to the manifold learning approach by \textcite{calandra2016manifold}.

The means of the variational distribution were initialized based on standard Principal Component Analysis (PCA), and the latent variances were initialized to 0.1. Also, 20 points from the initial latent space were selected as inducing pseudo-inputs and were appropriately optimized during training.

Each scenario was evaluated following two approaches: a qualitative analysis of the learned two-dimensional latent space, a quantitative metric in which we took the known labels from each dataset and computed the predictive accuracy of the predicted classes of points in the latent space. In the latter, we used a five-fold cross-validated 1-nearest neighbor (1-NN). For the quantitative results, we also show the accuracy of the PCA projection for reference.

\subsubsection{Oil Flow Dataset}

The multiphase Oil flow dataset consists of 1000 observations with 12 attributes, belonging to one of three classes \parencite{bishop1993}. We applied GPLVM with five latent dimensions and selected the two dimensions with the greatest inverse lengthscales.

For the approximations with the GH quadrature, we used $2^5=32$ samples. This contrasts with the UT, which only uses $2\cdot 5=10$ samples. Note that we have attempted to follow \textcite{titsias2010} and use ten latent dimensions, but that would require the GH to evaluate $2^{10}=1024$ samples at each optimization step, which made the method too slow on the tested hardware. 

\begin{figure*}[t]
    \begin{center}
    \begin{subfigure}[t]{.24\linewidth}
        \centering
        \includegraphics[width=\linewidth]{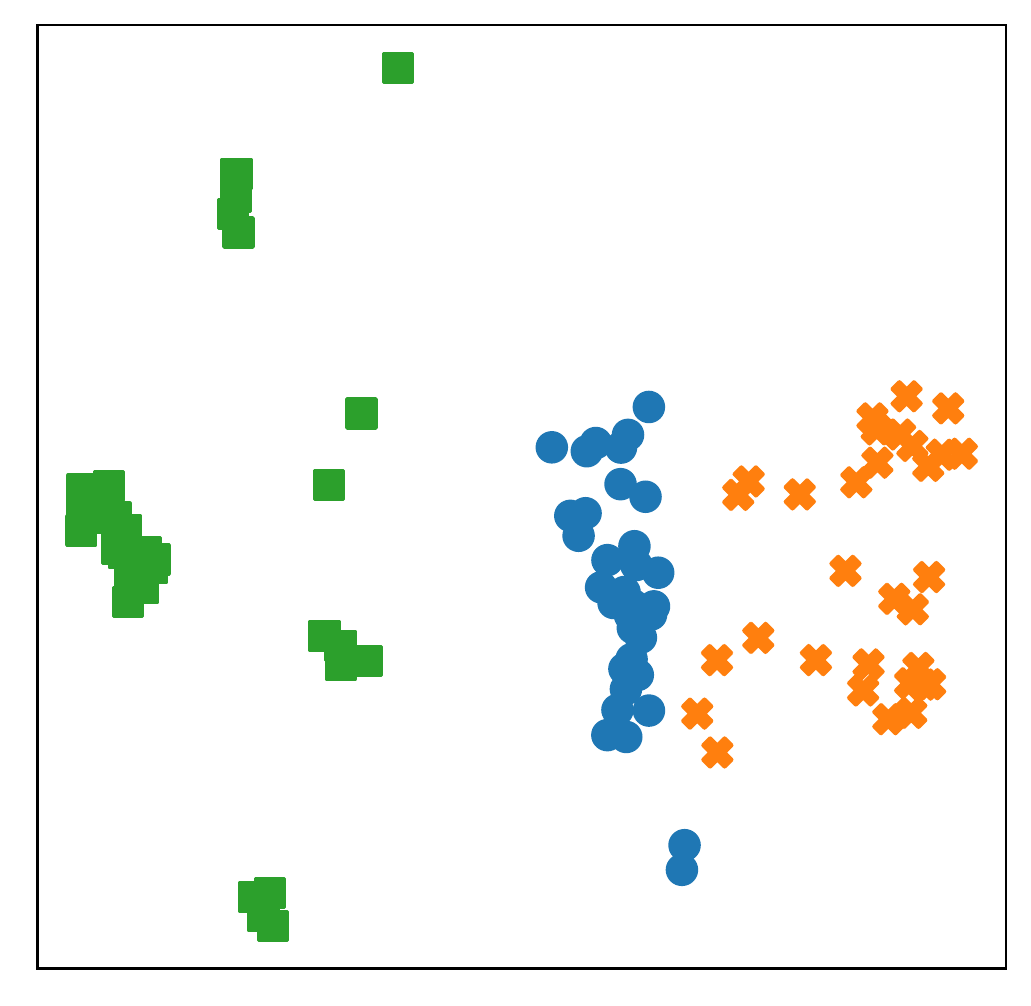}
        \caption{Analytic RBF.}
    \end{subfigure}
    \begin{subfigure}[t]{.24\linewidth}
        \centering
        \includegraphics[width=\linewidth]{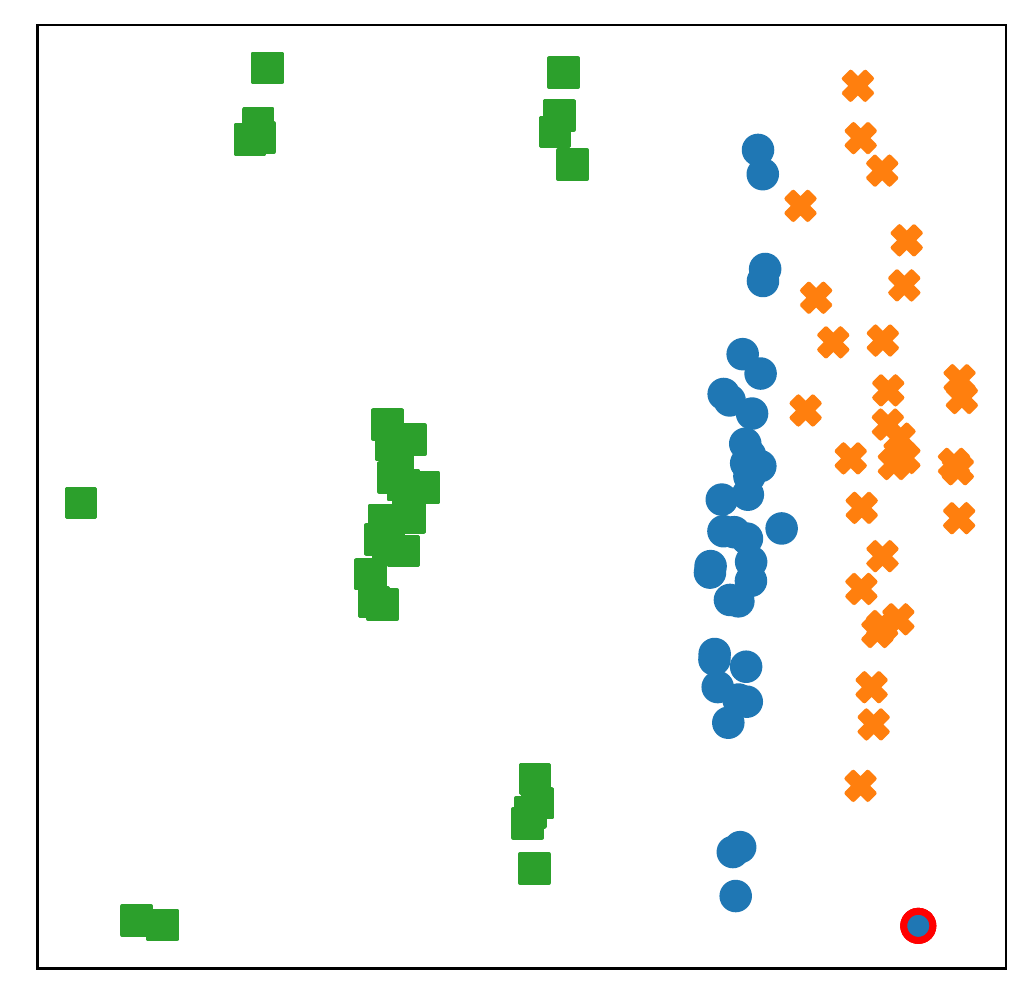}
        \caption{Matérn 3/2 (GH).}
    \end{subfigure}
    \begin{subfigure}[t]{.24\linewidth}
        \centering
        \includegraphics[width=\linewidth]{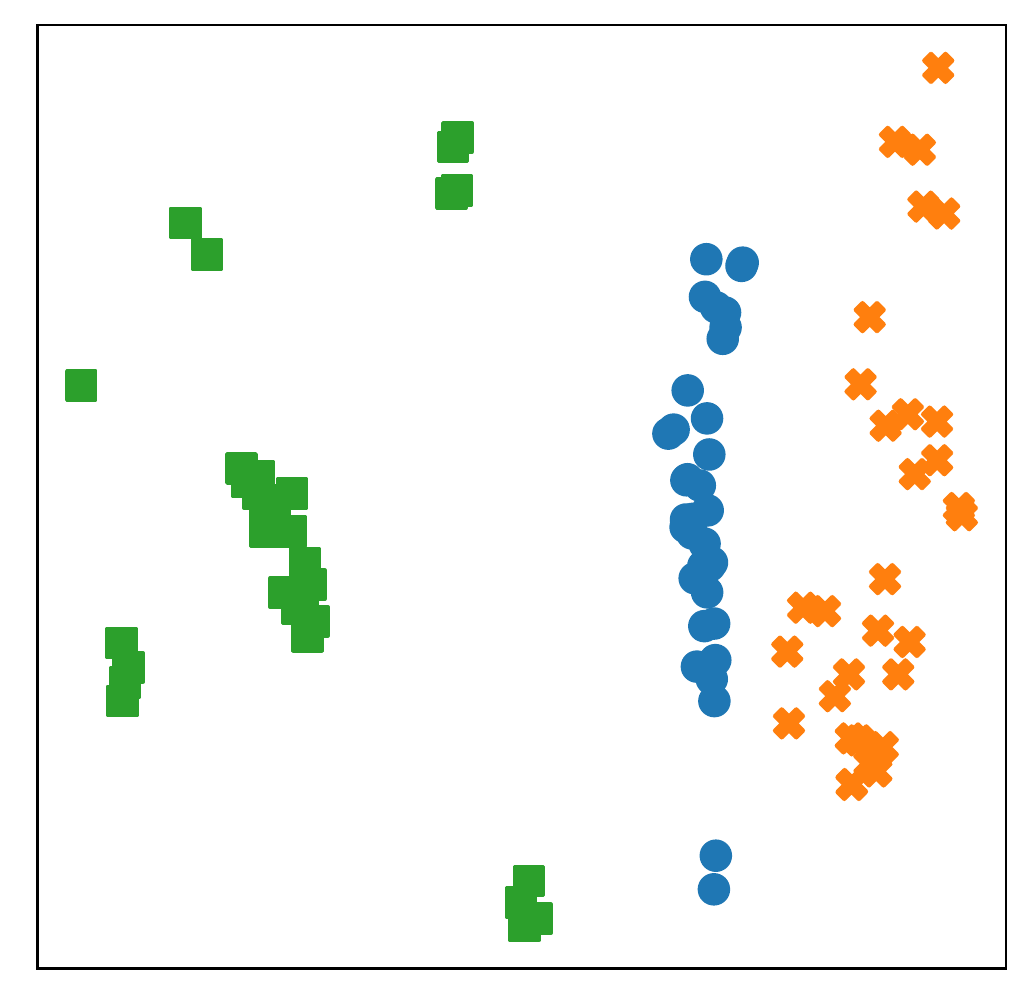}
        \caption{Matérn 3/2 (UT).}
    \end{subfigure}
    \begin{subfigure}[t]{.24\linewidth}
        \centering
        \includegraphics[width=\linewidth]{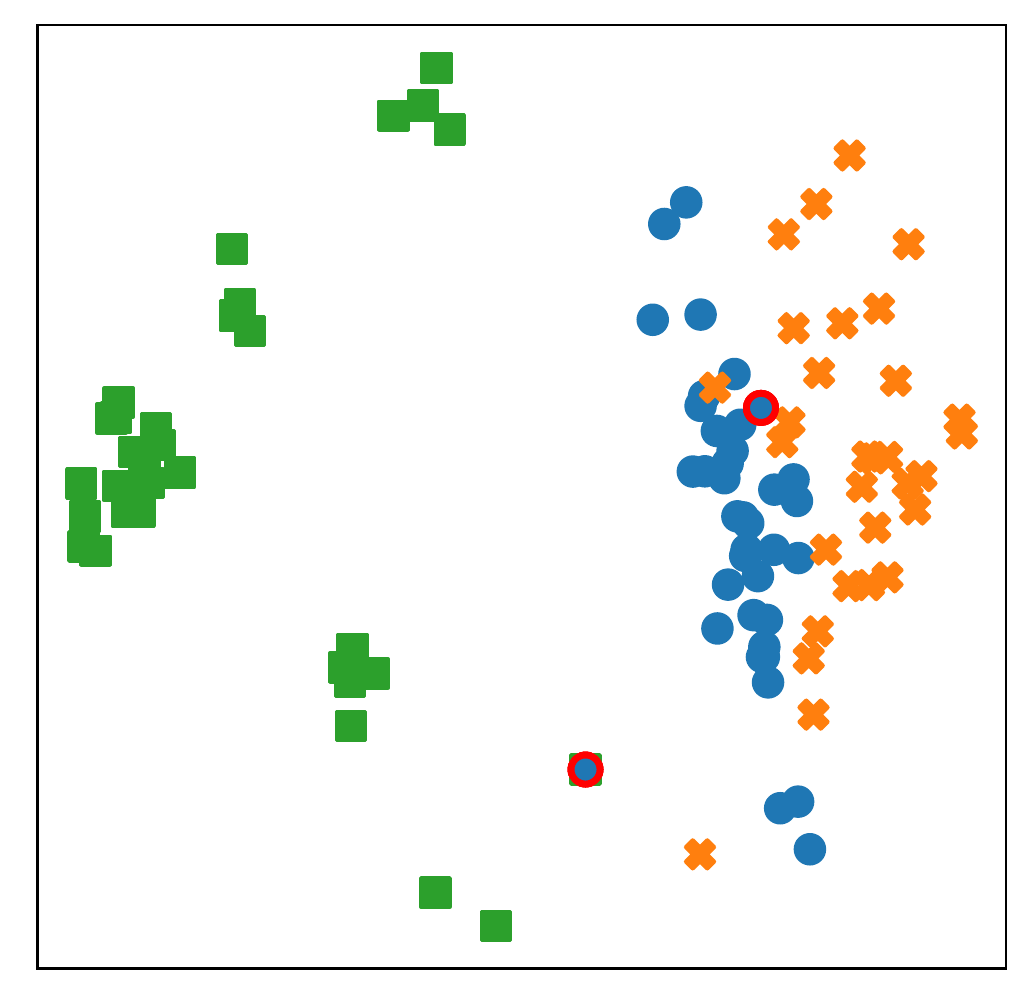}
        \caption{Matérn 3/2 (MC(32)).}
    \end{subfigure}
    \vspace{1ex}
    \hrule
    \vspace{1ex}
    \begin{subfigure}[t]{.24\linewidth}
        \centering
        \includegraphics[width=\linewidth]{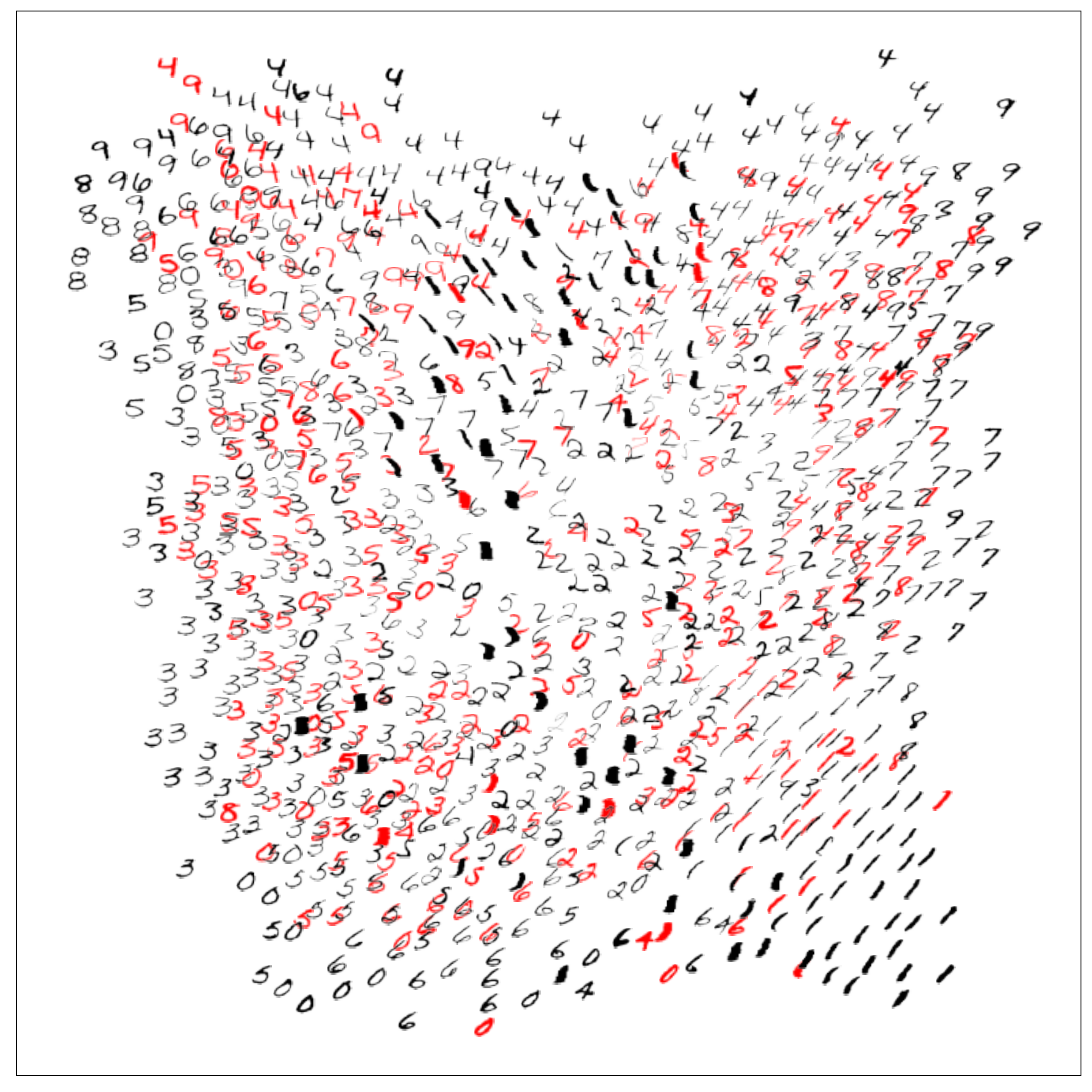}
        \caption{Analytic RBF.}
    \end{subfigure}
    \begin{subfigure}[t]{.24\linewidth}
        \centering
        \includegraphics[width=\linewidth]{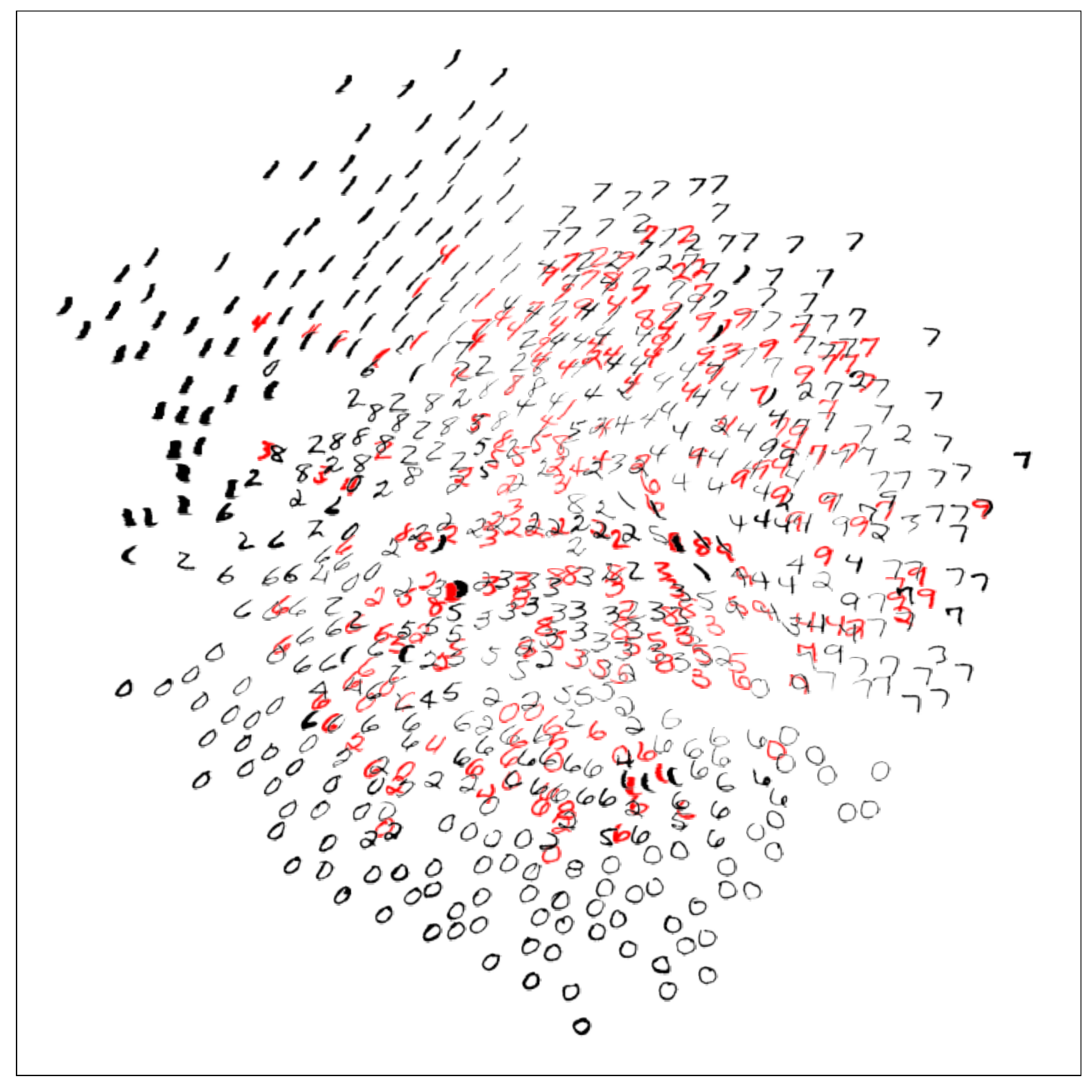}
        \caption{MLP kernel (GH).}
    \end{subfigure}
    \begin{subfigure}[t]{.24\linewidth}
        \centering
        \includegraphics[width=\linewidth]{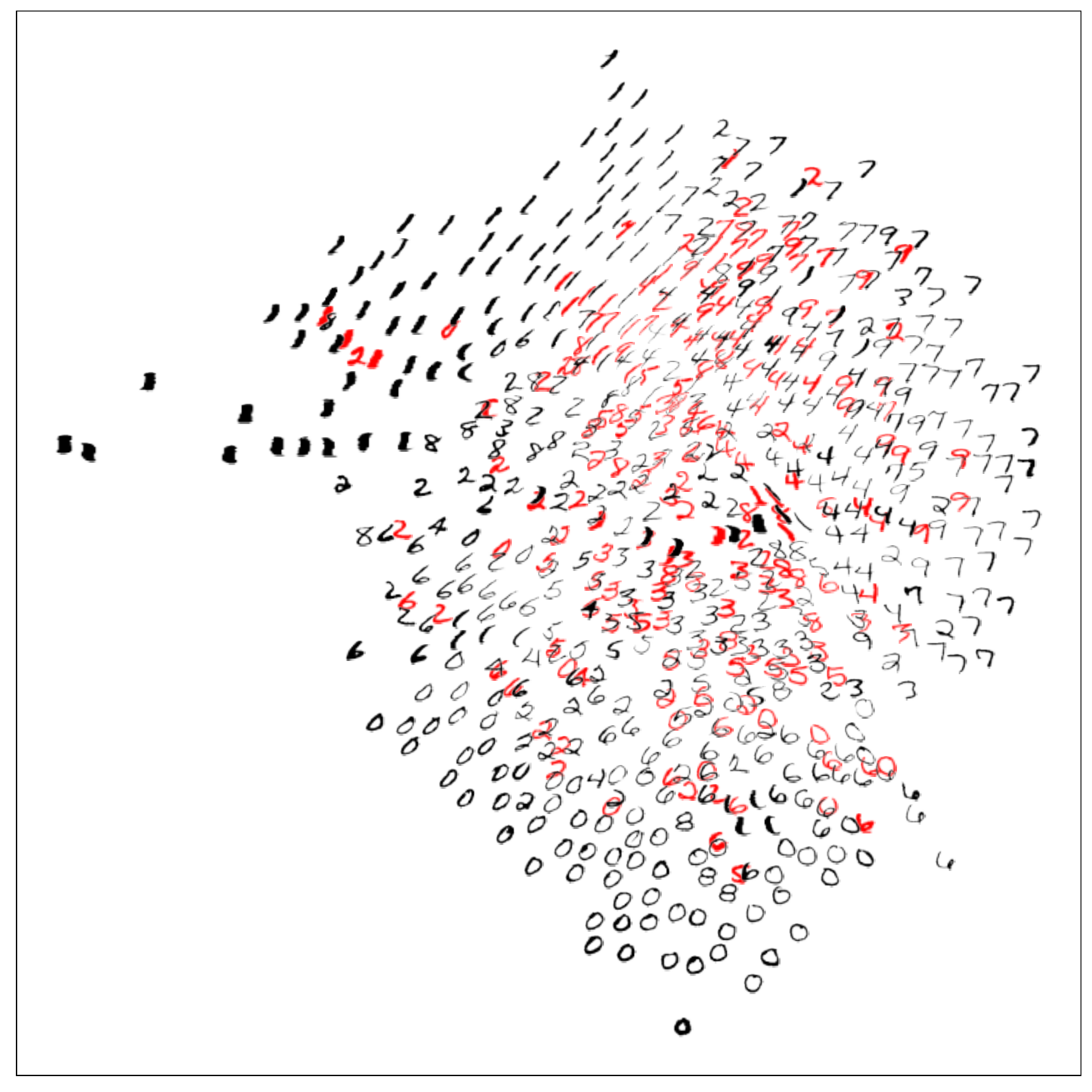}
        \caption{MLP kernel (UT).}
    \end{subfigure}
    \begin{subfigure}[t]{.24\linewidth}
        \centering
        \includegraphics[width=\linewidth]{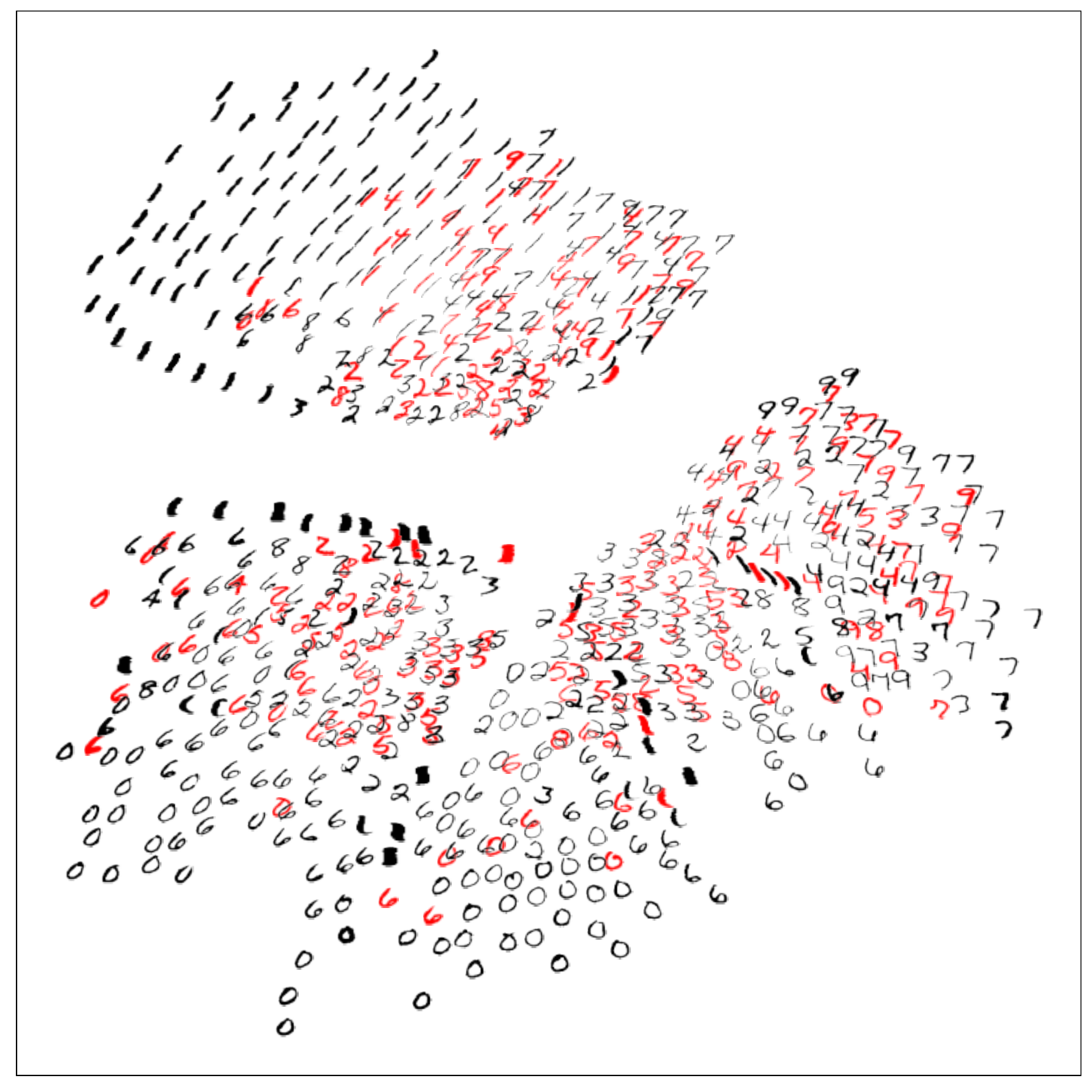}
        \caption{MLP kernel (MC(200)).}
    \end{subfigure}

    \end{center}
    \caption{Projections of the Oil flow and USPS digits datasets for GPLVM with different kernels and approximations. The projections shown are the ones with median score obtained in the cross-validation steps. 1-NN mislabels are marked in red. By visual inspection, MC approximations deviated the most from the other approximations despite having the same model as the others.}
    \label{fig:oilRBF}
\end{figure*}

\begin{table}[t]
    \caption{1-NN accuracies results for the Oil flow dataset. Note that the UT managed to achieve better results while using $\frac{1}{3}$ of the evaluations as GH.}
    \label{tab:dimred_oil}
    \centering
    \resizebox{\columnwidth}!{
    \begin{tabular}{
	    lcl
	    S[table-format=3.1(3),table-align-uncertainty=true]
	}
	\toprule
    \textbf{Method} & \textbf{\# evaluations} & \textbf{Kernel} & {\textbf{Accuracy}}\\
    \midrule
    PCA  &                    -  & -               & 79.0(65)  \\[1ex]
    Analytic        &                    -  & RBF  & 98.0(27)  \\[1ex]
    Gauss-Hermite   &                    32 & Matérn 3/2      & 95.0(61)  \\
                    &                       & RBF  & 98.0(27)  \\[1ex]
    Unscented       &                    10 & Matérn 3/2      & 100.0(0)  \\
                    &                       & RBF  & 98.0(27)  \\[1ex]
    Monte Carlo     &                    10 & Matérn 3/2      & 85.6(87) \\
                    &                       & RBF  & 98.2(24) \\
                    &                    32 & Matérn 3/2      & 87.9(54)  \\
                    &                       & RBF  & 98.0(25)  \\
                    &                   200 & Matérn 3/2      & 95.4(30)  \\
                    &                       & RBF  & 97.0(40)  \\
	\bottomrule
	\end{tabular}
 	}
\end{table}

In Figure \ref{fig:oilRBF}, it can be seen that independent of the chosen method to solve the $\Psi$-statistics, either the analytic expressions or any of the deterministic approximations yields similar overall qualitative results. Table \ref{tab:dimred_oil} contains the 1-NN predicted accuracy results for all kernels and approximation methods. As expected, all the nonlinear approaches performed better than regular PCA. The RBF results for the deterministic approaches are identical, while the Matérn 3/2 kernel with the UT approximation obtained slightly better results overall. However, when using MC estimates with the same amount of points that UT and GH used and the Matérn 3/2 kernel, the results were worse than both UT and GH.

\subsubsection{USPS Digit Dataset}

The USPS digit dataset contains 7000 $16\times 16$ gray-scale images of handwritten numerals from 0 to 9. To soften the required computational effort, we used just 500 samples of each class. We used a GPLVM with five latent dimensions on all kernels except the MLP kernel, where two latent dimensions were used. The same evaluation methodology previously described was followed.

We expected the MLP kernel to fare better than the RBF kernel due to neural networks' well-known capabilities to find lower-dimensional representations of higher dimensional structured data \parencite{wilson2016deep}. From Table \ref{tab:dimred_usps}, this was the case since all methods had an increase of 30\% accuracy compared to their results with RBF. We also noted that even MC approximations with more evaluations than UT and GH do not achieve the same results.

Figure \ref{fig:oilRBF} compares the analytic solution with RBF versus the approximate solutions using the MLP kernel with a single hidden layer and [2, 30, 60] neurons (input, hidden, and output, respectively). Visually, the difference between the kernels is as stark, as noted in the quantitative results. These plots also show that the MC approximation finds a very different projection than the other methods that are arguably more difficult to interpret due to the appearance of a gap in the latent data.

\begin{table}[h]
    \caption{1-NN accuracies results for the USPS dataset. The use of a more complex kernel brought benefits to all methods. Despite its simplicity, the UT has better or similar results on all kernels.}
    \label{tab:dimred_usps}
    \centering
    \resizebox{\columnwidth}!{
    \begin{tabular}{
	    lcl
	    S[table-format=2.1(2),table-align-uncertainty=true]
	}
	\toprule
    \textbf{Method} & \textbf{\# evaluations} & \textbf{Kernel} & {\textbf{Accuracy}}\\
    \midrule
    PCA  &                     - &                 &     35.5(14)\\[1ex]
    Analytic        &                     - & RBF  &     36.0(10)\\[1ex]
    Gauss-Hermite   &                     4 & MLP  &     68.8(9)\\
                    &                    32 & RBF  &     36.7(6)\\[1ex]
    Unscented       &                     4 & MLP  &     69.0(19)\\
                    &                    10 & RBF  &     39.0(12)\\[1ex]
    Monte Carlo     &                     4 & MLP  &     47.9(18)\\
                    &                    10 & RBF  &     26.6(14)\\
                    &                    32 & RBF  &     27.0(14)\\
                    &                   200 & MLP  &     54.3(17)\\
                    &                       & RBF  &     29.5(15)\\
	\bottomrule
	\end{tabular}
 	}
\end{table}

\subsection{Dynamical Free Simulation}

Free simulation, or multistep-ahead prediction, is a task that consists of forecasting the values of a dynamical system arbitrarily far into the future based on past predicted values. In most simple models, such as the GP-NARX \parencite{kocijan2005dynamic}, each prediction does not depend on the uncertainty of past predictions, but only past mean predicted values. The lack of dependency between the current predictions and the uncertainty of past predictions can be a significant problem because the user cannot be confident about the quality of the prediction if it does not consider the compounded errors from past estimates.

To propagate the uncertainty of each prediction to the next implies to perform predictions with uncertain inputs. This task has been tackled before, for instance, by \textcite{girard2003}, but for GP models using the RBF kernel.

In this section, we first trained a GP-NARX without considering uncertain inputs, following the regular NARX approach \parencite{kocijan2005dynamic}. Then, we applied the same optimized kernel hyperparameters in a GPLVM, selecting all the training inputs as pseudo-inputs. Finally, the GPLVM is used to perform a free simulation with uncertain inputs formed by the past predictive distributions. Since we applied approximations for computing the $\Psi$-statistics in the predictions, any valid kernel function can be chosen.

\subsubsection{Airline Passenger Dataset}

\begin{figure*}[ht!]
    \begin{center}
    \hspace*{\fill}
    \begin{subfigure}[b]{0.43\linewidth}
        \centering
        \includegraphics[width=\linewidth]{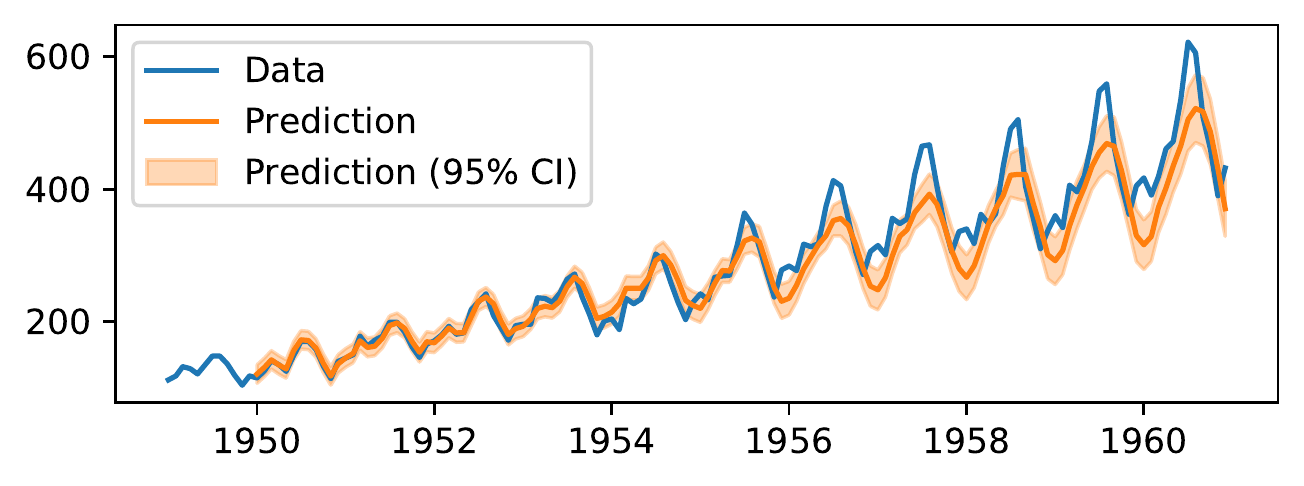}
        \caption{GP-NARX, Periodic + RBF + Linear.}
    \end{subfigure}
    \hfill
    \begin{subfigure}[b]{0.43\linewidth}
        \centering
        \includegraphics[width=\linewidth]{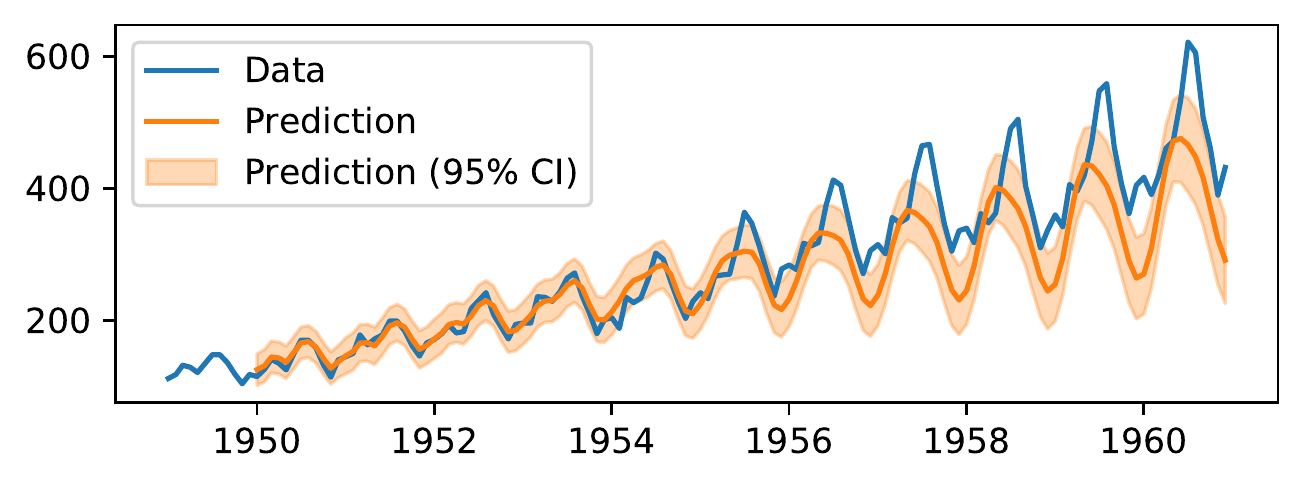}
        \caption{GPLVM, RBF + Linear.}
    \end{subfigure}
    \hspace*{\fill}
    \newline
    \hspace*{\fill}
    \begin{subfigure}[b]{0.43\linewidth}
        \centering
        \includegraphics[width=\linewidth]{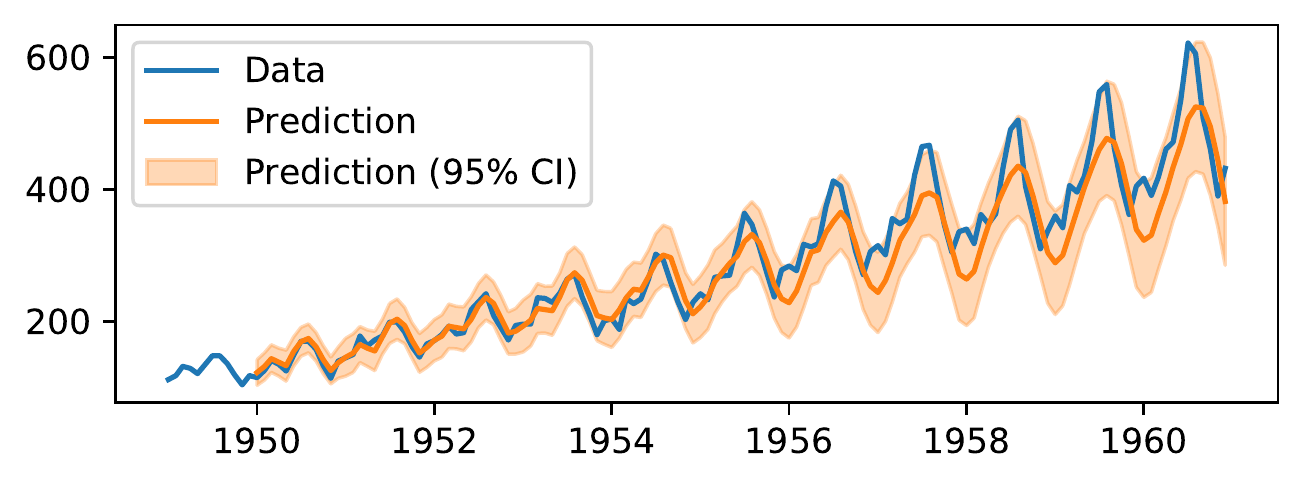}
        \caption{GPLVM, Periodic + RBF + Linear (GH).}
    \end{subfigure}
    \hfill
    \begin{subfigure}[b]{0.43\linewidth}
        \centering
        \includegraphics[width=\linewidth]{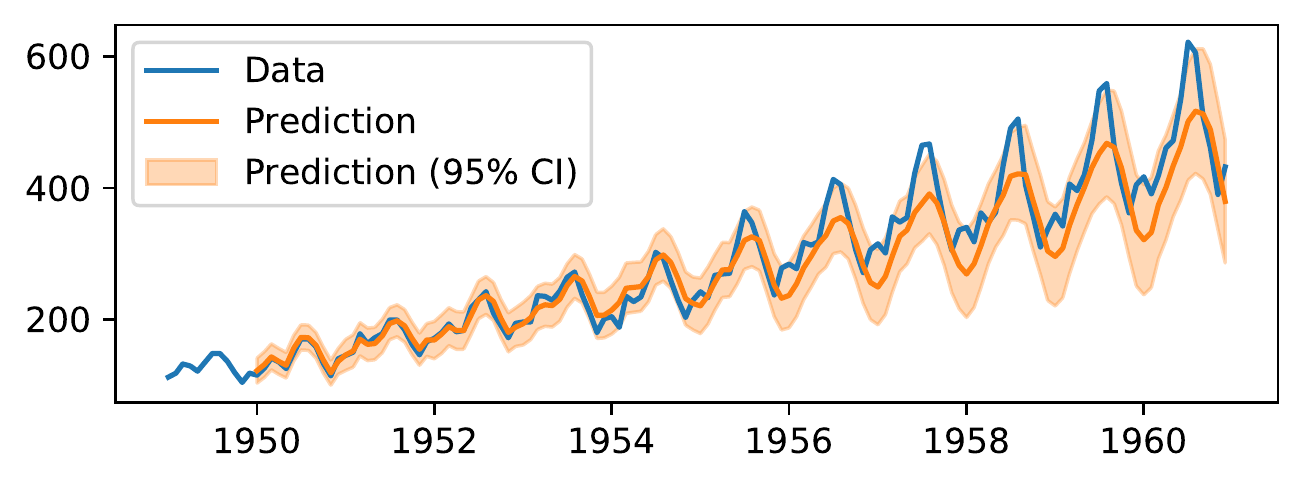}
        \caption{GPLVM, Periodic + RBF + Linear (UT).}
    \end{subfigure}
    \hspace*{\fill}
    \newline
    \hspace*{\fill}
    \begin{subfigure}[b]{0.43\linewidth}
        \centering
        \includegraphics[width=\linewidth]{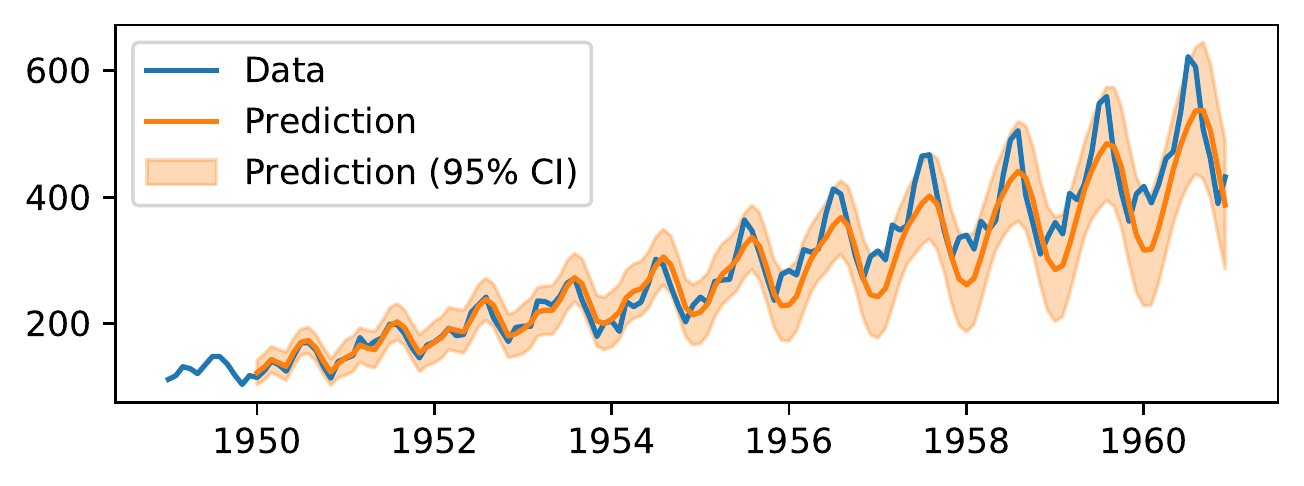}
        \caption{GPLVM, Periodic + RBF + Linear (MC(4096)).}
    \end{subfigure}
    \hfill
    \begin{subfigure}[b]{0.43\linewidth}
        \centering
        \includegraphics[width=\linewidth]{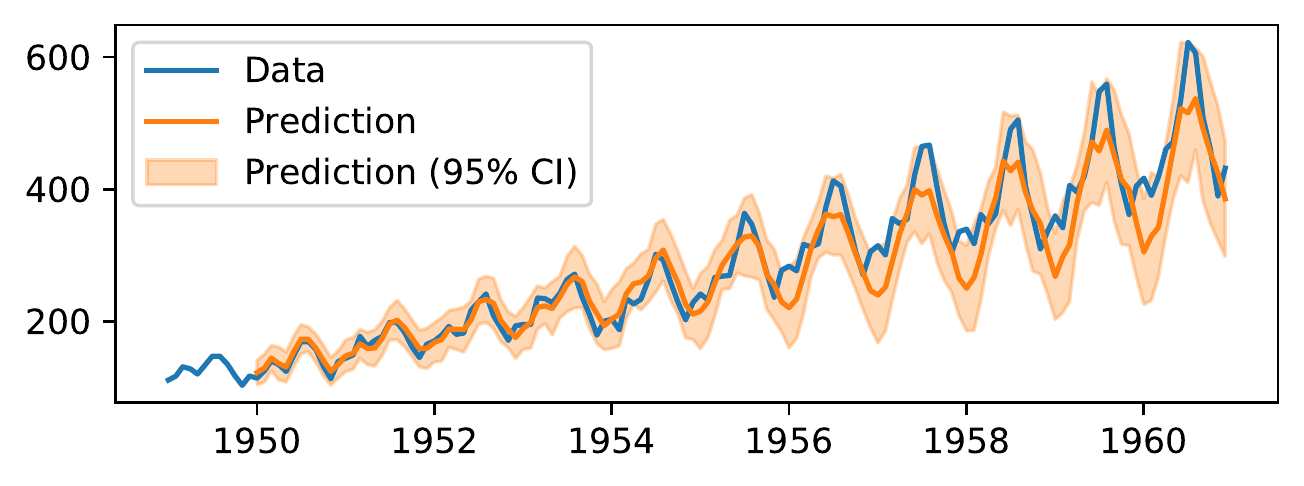}
        \caption{GPLVM, Periodic + RBF + Linear (MC(24)).}
    \end{subfigure}
    \hspace*{\fill}
    \end{center}
    \caption{Results obtained in the dynamical free simulation experiments. Best obtained runs are shown. Visibly, the MC approximation with 24 points has a much lower quality in its mean compared to the UT approximation by failing to model the peaks of the curve properly.}
    \label{fig:air}
\end{figure*}

The Airline passenger numbers dataset records monthly passenger numbers from 1949 to 1961 \parencite{timeseriesdata}. We used the first four years for training and left the rest for testing. We chose an autoregressive lag of 12 past observations as input. After the GP-NARX kernel hyperparameters are optimized, as previously mentioned, we choose the variance of the variational distribution in the GPLVM to be equal to the optimized noise variance. The free simulation starts from the beginning of the training set until the end of the test set, using past predicted variances as variational variances of the uncertain inputs, enabling approximate uncertainty propagation during the simulation.

We used the following kernels: a mixture of an RBF kernel with a linear kernel, a mixture of periodic\footnote{As defined by \textcite{mackay1998introduction} at Eq. (47).}, RBF, and linear kernels. The latter combination of kernels was chosen because of my prior knowledge that airplane ticket sales follow a periodic trend and have an overall upward tendency because of the popularity increase and decrease in ticket prices. We emphasize that the choice of such a flexible combination of kernels would not be possible without using approximate methods when considering the uncertain inputs scenario and the GPLVM framework.

Quantitative evaluation is done by computing the RMSE, given by
$\sqrt{\frac{1}{n^*} \sum_i^{n^*} (y_i - \mu_i^*)^2}$, where $n^*$ is the number of test samples, $y_i$ is the true output and $\mu_i^*$ is the predicted mean output. The average NLPD is also used as a evaluation metric, the NLPD score is given by $\frac{1}{2}\log2\pi + \frac{1}{2n^*} \sum_i^{n^*} \left[ \log{\sigma_i^*}^2 + \frac{(y_i - \mu_i^*)^2}{{\sigma_i^*}^2} \right],$ where ${\sigma_i^*}^2$ is the $i$-th predicted variance. Both metrics are ``the lower, the better'' and are computed only for the test set.

Table \ref{tab:airAcc} presents the obtained results. Although with similar RMSE, all GPLVM variants presented better NLPD values than their standard GP-NARX counterparts. That is expected since the uncertainty of each prediction is being approximately propagated to the next predictions. As for the models with UT, its results were better than the equivalent MC sample sizes but had a much better cost-benefit over the other methods given that they are using 8 to 170 times more samples for a 0.07 to 0.06 decrease in NLPD. As shown in Figure \ref{fig:air}, the visual difference between the two methods is subtle.

\begin{table}[ht]
    \caption{Summary of the free simulation results for the Air passengers dataset. All methods have comparable RMSE but when comparing with GH, a 170 fold increase in evaluations resulted with just a 0.06 decrease in NLPD.}
    \label{tab:airAcc}
    \centering
    \resizebox{\columnwidth}!{
    \begin{tabular}{
	    lcl
	    S[table-format=2.2(3),table-align-uncertainty=true]
	    S[table-format=2.2(3),table-align-uncertainty=true]
	}
	\toprule
	\textbf{Method} & \textbf{\# evaluations} & \textbf{Kernel} & \textbf{NLPD} & \textbf{RMSE} \\
	\midrule
	GP-NARX & -             & RBF+Linear                & 11.37    & 69.40\\
	        & -             & Per.+RBF+Lin.             &  7.46    & 44.98\\[1ex]

	GPLVM - Analytic & -    & RBF+Linear                &  7.08    & 68.93\\[1ex]

	GPLVM - GH & 4096
	                        & RBF+Linear                &  7.07    & 68.88\\
	        &               & Per.+RBF+Lin.             &  5.20    & 45.00\\[1ex]
	GPLVM - UT & 24
	                        & RBF+Linear                &  7.10    & 69.11\\
	        &               & Per.+RBF+Lin.             &  5.26    & 45.27\\[1ex]
	GPLVM - MC& 24
	                        & RBF+Linear                & 7.52(41) & 71.16(315)\\
	        &               & Per.+RBF+Lin.             & 5.41(17) & 46.99(304)\\[1ex]
	        &200
	                        & RBF+Linear                & 7.09(20) & 68.82(209)\\
	        &               & Per.+RBF+Lin.             & 5.19(6)  & 45.19(132)\\[1ex]
	        &4096
	                        & RBF+Linear                & 7.07(3)  & 68.81(37)\\
	        &               & Per.+RBF+Lin.             & 5.19(1)  & 45.29(28)\\[1ex]
	\bottomrule
	\end{tabular}
 	}
\end{table}

\section{RELATED WORK}
\label{SEC_REL}
A few authors have already considered the UT in the context of GP models. For instance, \textcite{ko2007gp,ko2009gp} propose using the Unscented Kalman Filter (UKF) with GP-based transition and observation functions, and others have successfully applied the resulting GP-UKF \parencite{anger2012unscented,wang2014human,safarinejadian2014fault}. \textcite{ko2011learning} extend the previous works by considering the original GPLVM \parencite{lawrence2004}, where the latent variables are optimized, instead of integrated.%
\textcite{steinberg2014extended} tackle other kinds of intractabilities and use the UT in GP models with non-Gaussian likelihoods in a variational framework. The resulting Unscented Gaussian Processes (UGP) is evaluated in synthetic inversion problems and binary classification. Later, \textcite{bonilla16} generalize that methodology to solve multi-output and multi-task problems while also enabling non-Gaussian likelihoods.

In summary, the GP-UKF and related models use GPs for filtering by basing their models on unscented Kalman filters. Furthermore, the UGP and related models focus on using the UT to solve intractable integrals that arise when considering non-Gaussian likelihoods in GP models. However, this paper's subject matter is the use of arbitrary kernels through the UT in Bayesian GPLVM models, where the latent variables which represent uncertain inputs are approximately marginalized.

\section{CONCLUSION}
\label{SEC_CONC}

In this paper, we considered learning GP models from unavailable or uncertain inputs within the Bayesian GPLVM framework. We tackled the intractabilities that arise in the original variational methodology by \textcite{titsias2010} when non-RBF and nonlinear kernels are used by proposing the use of the unscented transformation.

We performed experiments on two tasks: dimensionality reduction and free simulation of dynamical models with uncertainty propagation. In both cases, the UT-based approach scaled much better than the compared Gauss-Hermite quadrature, while obtaining a similar overall approximation in our experiments. The UT results were also more stable and consistent than those obtained by Monte Carlo sampling, which may require a more significant number of samples and can not be used with the popular quasi-Newton BFGS optimization algorithm. Importantly, the method is simple to implement and does not impose any stochasticity, maintaining the deterministic inference feature of the standard Bayesian GPLVM variational framework.

For future work, we aim to evaluate how other methods of obtaining sigma points might increase or decrease the quality of the approximations taken. Also, we intend to evaluate the UT in more scenarios where inference with GP models falls into intractable expectations. For instance, we intend to tackle integrals that arise with DGP models that have intractable inference due to low-dimensional integrals, like the doubly stochastic Gaussian process by \textcite{salimbeni2017doubly} and recurrent Gaussian processes by \textcite{mattos2016_iclr}.

\subsubsection*{References}
\printbibliography[heading=none]

@InProceedings{girard2003,
  author    = {Girard, Agathe and Rasmussen, Carl Edward and Candela, Joaquin Qui\~nonero and Murray-Smith, Roderick},
  booktitle = {Advances in Neural Information Processing Systems 15},
  title     = {{G}aussian process priors with uncertain inputs application to multiple-step ahead time series forecasting},
  year      = {2003},
  editor    = {S. Becker and S. Thrun and K. Obermayer},
  pages     = {545--552},
  publisher = {MIT Press},
  url       = {http://papers.nips.cc/paper/2313-gaussian-process-priors-with-uncertain-inputs-application-to-multiple-step-ahead-time-series-forecasting.pdf},
}

@InCollection{mackay1998introduction,
  author    = {D. J. C. MacKay},
  booktitle = {Neural Networks and Machine Learning},
  publisher = {Kluwer Academic Press},
  title     = {Introduction to {Gaussian} Processes},
  year      = {1998},
  editor    = {C. M. Bishop},
  pages     = {133--166},
  series    = {NATO ASI Series},
}

@InProceedings{lawrence2004,
  author    = {Neil D. Lawrence},
  booktitle = {Advances in Neural Information Processing Systems 16},
  title     = {{Gaussian} Process Latent Variable Models for Visualisation of High Dimensional Data},
  year      = {2004},
  editor    = {S. Thrun and L. K. Saul and B. Sch{\"{o}}lkopf},
  pages     = {329--336},
  publisher = {MIT Press},
  url       = {http://papers.nips.cc/paper/2540-gaussian-process-latent-variable-models-for-visualisation-of-high-dimensional-data.pdf},
}

@InProceedings{titsias2010,
  author    = {Michalis K. Titsias and Neil D. Lawrence},
  booktitle = {Proceedings of the Thirteenth International Conference on Artificial Intelligence and Statistics},
  title     = {{Bayesian} {Gaussian} Process Latent Variable Model},
  year      = {2010},
  address   = {Chia Laguna Resort, Sardinia, Italy},
  editor    = {Yee Whye Teh and Mike Titterington},
  pages     = {844--851},
  publisher = {PMLR},
  series    = {Proceedings of Machine Learning Research},
  volume    = {9},
  url       = {http://proceedings.mlr.press/v9/titsias10a.html},
}

@Book{rasmussen2006,
  author    = {Rasmussen, Carl and Williams, Chris},
  publisher = {MIT Press},
  title     = {{Gaussian} Processes for Machine Learning},
  year      = {2006},
  address   = {Cambridge, MA, USA},
  edition   = {1},
  isbn      = {9780262182539},
}

@InProceedings{steinberg2014extended,
  author    = {Steinberg, Daniel M and Bonilla, Edwin V},
  booktitle = {Advances in Neural Information Processing Systems 27},
  title     = {Extended and Unscented {Gaussian} Processes},
  year      = {2014},
  editor    = {Z. Ghahramani and M. Welling and C. Cortes and N. D. Lawrence and K. Q. Weinberger},
  pages     = {1251--1259},
  publisher = {Curran Associates, Inc.},
  url       = {http://papers.nips.cc/paper/5455-extended-and-unscented-gaussian-processes.pdf},
}

@Article{damianou2016variational,
  author  = {Andreas C. Damianou and Michalis K. Titsias and Neil D. Lawrence},
  journal = {Journal of Machine Learning Research},
  title   = {Variational inference for latent variables and uncertain inputs in {Gaussian} processes},
  year    = {2016},
  number  = {42},
  pages   = {1--62},
  volume  = {17},
  url     = {http://jmlr.org/papers/v17/damianou16a.html},
}

@InProceedings{wilson2013gaussian,
  author    = {Andrew Wilson and Ryan Adams},
  booktitle = {Proceedings of the 30th International Conference on Machine Learning},
  title     = {{Gaussian} Process Kernels for Pattern Discovery and Extrapolation},
  year      = {2013},
  address   = {Atlanta, GA, USA},
  editor    = {Sanjoy Dasgupta and David McAllester},
  number    = {3},
  pages     = {1067--1075},
  publisher = {PMLR},
  series    = {Proceedings of Machine Learning Research},
  volume    = {28},
  url       = {http://proceedings.mlr.press/v28/wilson13.html},
}

@InProceedings{wilson2016deep,
  author    = {Andrew Gordon Wilson and Zhiting Hu and Ruslan Salakhutdinov and Eric P. Xing},
  booktitle = {Proceedings of the 19th International Conference on Artificial Intelligence and Statistics},
  title     = {Deep Kernel Learning},
  year      = {2016},
  address   = {Cadiz, Spain},
  editor    = {Arthur Gretton and Christian C. Robert},
  pages     = {370--378},
  publisher = {PMLR},
  series    = {Proceedings of Machine Learning Research},
  volume    = {51},
  url       = {http://proceedings.mlr.press/v51/wilson16.html},
}

@InProceedings{wilson2016stochastic,
  title = {Stochastic Variational Deep Kernel Learning},
  author = {Wilson, Andrew G and Hu, Zhiting and Salakhutdinov, Russ R and Xing, Eric P},
  booktitle = {Advances in Neural Information Processing Systems 29},
  editor = {D. D. Lee and M. Sugiyama and U. V. Luxburg and I. Guyon and R. Garnett},
  pages = {2586--2594},
  year = {2016},
  publisher = {Curran Associates, Inc.},
  url = {http://papers.nips.cc/paper/6426-stochastic-variational-deep-kernel-learning.pdf}
}

@InProceedings{mattos2016_iclr,
  author        = {C{\'{e}}sar Lincoln C. Mattos and Zhenwen Dai and Andreas C. Damianou and Jeremy Forth and Guilherme A. Barreto and Neil D. Lawrence},
  booktitle     = {4th International Conference on Learning Representations, ICLR 2016},
  title         = {Recurrent {Gaussian} Processes},
  year          = {2016},
  editor        = {Yoshua Bengio and Yann LeCun},
  archiveprefix = {arxiv},
  bibsource     = {dblp computer science bibliography, https://dblp.org},
  eprint        = {1511.06644},
}

@InProceedings{damianou2013,
  author    = {Andreas Damianou and Neil Lawrence},
  booktitle = {Proceedings of the Sixteenth International Conference on Artificial Intelligence and Statistics},
  title     = {Deep {Gaussian} processes},
  year      = {2013},
  address   = {Scottsdale, AZ, USA},
  editor    = {Carlos M. Carvalho and Pradeep Ravikumar},
  pages     = {207--215},
  publisher = {PMLR},
  series    = {Proceedings of Machine Learning Research},
  volume    = {31},
  url       = {http://proceedings.mlr.press/v31/damianou13a.html},
}

@InProceedings{casale2018gaussian,
  author    = {Casale, Francesco Paolo and Dalca, Adrian and Saglietti, Luca and Listgarten, Jennifer and Fusi, Nicolo},
  booktitle = {Advances in Neural Information Processing Systems 31},
  date      = {2018},
  title     = {{Gaussian} Process Prior Variational Autoencoders},
  editor    = {S. Bengio and H. Wallach and H. Larochelle and K. Grauman and N. Cesa-Bianchi and R. Garnett},
  pages     = {10369--10380},
  publisher = {Curran Associates, Inc.},
  url       = {http://papers.nips.cc/paper/8238-gaussian-process-prior-variational-autoencoders.pdf},
}

@InProceedings{havasi2018inference,
  author    = {Havasi, Marton and Hernandez-Lobato, Jose Miguel and Murillo-Fuentes, Juan Jos\'e},
  booktitle = {Advances in Neural Information Processing Systems 31},
  date      = {2018},
  title     = {Inference in Deep {Gaussian} Processes using Stochastic Gradient Hamiltonian Monte Carlo},
  editor    = {S. Bengio and H. Wallach and H. Larochelle and K. Grauman and N. Cesa-Bianchi and R. Garnett},
  pages     = {7506--7516},
  publisher = {Curran Associates, Inc.},
  url       = {http://papers.nips.cc/paper/7979-inference-in-deep-gaussian-processes-using-stochastic-gradient-hamiltonian-monte-carlo.pdf},
}

@InProceedings{duvenaud2013structure,
  author    = {David Duvenaud and James Lloyd and Roger Grosse and Joshua Tenenbaum and Ghahramani Zoubin},
  booktitle = {Proceedings of the 30th International Conference on Machine Learning},
  title     = {Structure discovery in nonparametric regression through compositional kernel search},
  year      = {2013},
  address   = {Atlanta, GA, USA},
  editor    = {Sanjoy Dasgupta and David McAllester},
  number    = {3},
  pages     = {1166--1174},
  publisher = {PMLR},
  series    = {Proceedings of Machine Learning Research},
  volume    = {28},
  url       = {http://proceedings.mlr.press/v28/duvenaud13.html},
}

@InProceedings{lloyd2014automatic,
  author    = {Lloyd, James Robert and Duvenaud, David and Grosse, Roger and Tenenbaum, Joshua B. and Ghahramani, Zoubin},
  booktitle = {Proceedings of the Twenty-Eighth AAAI Conference on Artificial Intelligence},
  title     = {Automatic Construction and Natural-Language Description of Nonparametric Regression Models},
  year      = {2014},
  address   = {Qu{\'{e}}bec City, Qu{\'{e}}bec, Canada},
  pages     = {1242--1250},
  publisher = {AAAI Press},
  series    = {AAAI{\textquoteright}14},
  numpages  = {9},
}

@Article{bishop1993,
  author    = {Christopher M. Bishop and Gavin D. James},
  journal   = {Nuclear Instruments and Methods in Physics Research Section A: Accelerators, Spectrometers, Detectors and Associated Equipment},
  title     = {Analysis of multiphase flows using dual-energy gamma densitometry and neural networks},
  year      = {1993},
  number    = {2-3},
  pages     = {580--593},
  volume    = {327},
  doi       = {10.1016/0168-9002(93)90728-Z},
  publisher = {Elsevier BV},
}

@InProceedings{eleftheriadis2017identification,
  author    = {Eleftheriadis, Stefanos and Nicholson, Tom and Deisenroth, Marc and Hensman, James},
  booktitle = {Advances in Neural Information Processing Systems 30},
  title     = {Identification of {Gaussian} process state space models},
  year      = {2017},
  editor    = {I. Guyon and U. V. Luxburg and S. Bengio and H. Wallach and R. Fergus and S. Vishwanathan and R. Garnett},
  pages     = {5309--5319},
  publisher = {Curran Associates, Inc.},
  url       = {http://papers.nips.cc/paper/7115-identification-of-gaussian-process-state-space-models.pdf},
}

@InCollection{eleftheriadis2016variational,
  author    = {Stefanos Eleftheriadis and Ognjen Rudovic and Marc Peter Deisenroth and Maja Pantic},
  booktitle = {Computer Vision - ACCV 2016},
  date      = {2017},
  title     = {Variational {Gaussian} Process Auto-Encoder for Ordinal Prediction of Facial Action Units},
  doi       = {10.1007/978-3-319-54184-6_10},
  editor    = {Lai, Shang-Hong and Lepetit, Vincent and Nishino, Ko and Sato, Yoichi},
  isbn      = {978-3-319-54184-6},
  location  = {Cham},
  pages     = {154--170},
  publisher = {Springer International Publishing},
}

@InCollection{salimbeni2017doubly,
  author    = {Salimbeni, Hugh and Deisenroth, Marc},
  booktitle = {Advances in Neural Information Processing Systems 30},
  publisher = {Curran Associates, Inc.},
  title     = {Doubly Stochastic Variational Inference for Deep {Gaussian} Processes},
  year      = {2017},
  editor    = {I. Guyon and U. V. Luxburg and S. Bengio and H. Wallach and R. Fergus and S. Vishwanathan and R. Garnett},
  pages     = {4588--4599},
  url       = {http://papers.nips.cc/paper/7045-doubly-stochastic-variational-inference-for-deep-gaussian-processes.pdf},
}

@Article{al2017learning,
  author  = {Maruan Al-Shedivat and Andrew Gordon Wilson and Yunus Saatchi and Zhiting Hu and Eric P. Xing},
  journal = {Journal of Machine Learning Research},
  title   = {Learning scalable deep kernels with recurrent structure},
  year    = {2017},
  issn    = {1532-4435},
  number  = {82},
  pages   = {1--37},
  volume  = {18},
  editor  = {Murphy, Kevin and Sch{\"{o}}lkopf, Bernhard},
  url     = {http://jmlr.org/papers/v18/16-498.html},
}

@InProceedings{KingmaWelling2014,
  author    = {Diederik P. Kingma and Max Welling},
  booktitle = {2nd International Conference on Learning Representations, {ICLR} 2014},
  title     = {Auto-Encoding Variational {Bayes}},
  year      = {2014},
  address   = {Banff, AB, Canada},
  editor    = {Yoshua Bengio and Yann LeCun},
  eprint    = {1312.6114},
}

@InProceedings{RezendeEtAl2014,
  author    = {Danilo Jimenez Rezende and Shakir Mohamed and Daan Wierstra},
  booktitle = {Proceedings of the 31st International Conference on Machine Learning},
  title     = {Stochastic Backpropagation and Approximate Inference in Deep Generative Models},
  year      = {2014},
  address   = {Bejing, China},
  editor    = {Eric P. Xing and Tony Jebara},
  number    = {2},
  pages     = {1278--1286},
  publisher = {PMLR},
  series    = {Proceedings of Machine Learning Research},
  volume    = {32},
  url       = {http://proceedings.mlr.press/v32/rezende14.html},
}

@InProceedings{titsias2014doubly,
  author    = {Michalis K. Titsias and Miguel L{\'{a}}zaro-Gredilla},
  booktitle = {Proceedings of the 31st International Conference on International Conference on Machine Learning - Volume 32},
  title     = {Doubly Stochastic Variational {Bayes} for Non-Conjugate Inference},
  year      = {2014},
  address   = {Beijing, China},
  pages     = {II--1971--II--1980},
  publisher = {JMLR.org},
  series    = {ICML{\textquoteright}14},
  numpages  = {9},
}

@InProceedings{ko2007gp,
  author       = {Jonathan Ko and Daniel J. Klein and Dieter Fox and Dirk Haehnel},
  booktitle    = {2007 IEEE/RSJ International Conference on Intelligent Robots and Systems},
  title        = {{GP-UKF}: Unscented {Kalman} filters with {Gaussian} process prediction and observation models},
  year         = {2007},
  organization = {IEEE},
  pages        = {1901--1907},
  publisher    = {IEEE},
  doi          = {10.1109/iros.2007.4399284},
}

@Article{julier2004unscented,
  author    = {S.J. Julier and J.K. Uhlmann},
  journal   = {Proceedings of the IEEE},
  title     = {Unscented Filtering and Nonlinear Estimation},
  year      = {2004},
  number    = {3},
  pages     = {401--422},
  volume    = {92},
  doi       = {10.1109/jproc.2003.823141},
  publisher = {Institute of Electrical and Electronics Engineers (IEEE)},
}

@Article{menegaz2015systematization,
  author    = {Henrique M. T. Menegaz and Joao Y. Ishihara and Geovany A. Borges and Alessandro N. Vargas},
  journal   = {IEEE Transactions on Automatic Control},
  title     = {A Systematization of the Unscented {Kalman} Filter Theory},
  year      = {2015},
  number    = {10},
  pages     = {2583--2598},
  volume    = {60},
  doi       = {10.1109/tac.2015.2404511},
  publisher = {Institute of Electrical and Electronics Engineers (IEEE)},
}

@Article{ko2009gp,
  author    = {Jonathan Ko and Dieter Fox},
  journal   = {Autonomous Robots},
  title     = {{GP-BayesFilters}: {Bayesian} filtering using {Gaussian} process prediction and observation models},
  year      = {2009},
  number    = {1},
  pages     = {75--90},
  volume    = {27},
  doi       = {10.1007/s10514-009-9119-x},
  publisher = {Springer},
}

@Article{ko2011learning,
  author    = {Jonathan Ko and Dieter Fox},
  journal   = {Autonomous Robots},
  title     = {Learning {GP-BayesFilters} via {Gaussian} process latent variable models},
  year      = {2010},
  number    = {1},
  pages     = {3--23},
  volume    = {30},
  doi       = {10.1007/s10514-010-9213-0},
  publisher = {Springer},
}

@InProceedings{anger2012unscented,
  author    = {Christoph Anger and Robert Schrader and Uwe Klingauf},
  booktitle = {Proceedings of the european conference of the PHM society},
  title     = {Unscented {Kalman} filter with {Gaussian} process degradation model for bearing fault prognosis},
  year      = {2012},
  address   = {Dresden, Germany},
  editor    = {Anibal Bregon and Abhinav Saxena},
}

@InProceedings{wang2014human,
  author       = {Ziyou Wang and Jun Kinugawa and Hongbo Wang and Kosuge Kazahiro},
  booktitle    = {2014 IEEE International Conference on Information and Automation (ICIA)},
  title        = {A human motion estimation method based on {GP-UKF}},
  year         = {2014},
  address      = {Hailar, China},
  organization = {IEEE},
  pages        = {1228--1232},
  publisher    = {IEEE},
  doi          = {10.1109/icinfa.2014.6932836},
}

@Article{safarinejadian2014fault,
  author    = {Behrooz Safarinejadian and Elham Kowsari},
  journal   = {Systems Science \& Control Engineering},
  title     = {Fault detection in non-linear systems based on {GP-EKF} and {GP-UKF} algorithms},
  year      = {2014},
  number    = {1},
  pages     = {610--620},
  volume    = {2},
  doi       = {10.1080/21642583.2014.956843},
  publisher = {Informa UK Limited},
}

@InProceedings{titsias2009,
  author    = {Michalis K. Titsias},
  booktitle = {Proceedings of the Twelth International Conference on Artificial Intelligence and Statistics},
  title     = {Variational Learning of Inducing Variables in Sparse {Gaussian} Processes},
  year      = {2009},
  address   = {Hilton Clearwater Beach Resort, Clearwater Beach, Florida USA},
  editor    = {David van Dyk and Max Welling},
  pages     = {567--574},
  publisher = {PMLR},
  series    = {Proceedings of Machine Learning Research},
  volume    = {5},
  url       = {http://proceedings.mlr.press/v5/titsias09a.html},
}

@Article{jordan1999,
  author    = {Michael I. Jordan and Zoubin Ghahramani and Tommi S. Jaakkola and Lawrence K. Saul},
  journal   = {Machine Learning},
  title     = {An introduction to variational methods for graphical models},
  year      = {1999},
  number    = {2},
  pages     = {183--233},
  volume    = {37},
  doi       = {10.1023/a:1007665907178},
  publisher = {Springer},
}

@Misc{timeseriesdata,
  author = {Rob J Hyndman},
  title  = {Time Series Data Library},
  year   = {2018},
  url    = {https://datamarket.com/data/list/?q=provider:tsdl},
}

@PhdThesis{uhlmann1995dynamic,
  author      = {Jeffrey K. Uhlmann},
  school      = {University of Oxford},
  title       = {Dynamic map building and localization: New theoretical foundations},
  year        = {1995},
  institution = {University of Oxford},
  url         = {http://faculty.missouri.edu/uhlmannj/ThesisScan.pdf},
}

@Article{matthews2017gpflow,
  author  = {Alexander G. de G. Matthews and Mark van der Wilk and Tom Nickson and Keisuke Fujii and Alexis Boukouvalas and Pablo Le{\'{o}}n-Villagr{\'{a}} and Zoubin Ghahramani and James Hensman},
  journal = {Journal of Machine Learning Research},
  title   = {{GPflow}: A {Gaussian} Process Library using {TensorFlow}},
  year    = {2017},
  number  = {40},
  pages   = {1--6},
  volume  = {18},
  url     = {http://jmlr.org/papers/v18/16-537.html},
}

@InProceedings{calandra2016manifold,
  author       = {Roberto Calandra and Jan Peters and Carl Edward Rasmussen and Marc Peter Deisenroth},
  booktitle    = {2016 International Joint Conference on Neural Networks (IJCNN)},
  title        = {Manifold {Gaussian} processes for regression},
  year         = {2016},
  address      = {Vancouver, BC, Canada},
  organization = {IEEE},
  pages        = {3338--3345},
  publisher    = {IEEE},
  doi          = {10.1109/ijcnn.2016.7727626},
}

@Article{kocijan2005dynamic,
  author    = {Ju{\v{s}} Kocijan and Agathe Girard and Bla{\v{z}} Banko and Roderick Murray-Smith},
  journal   = {Mathematical and Computer Modelling of Dynamical Systems},
  title     = {Dynamic systems identification with {Gaussian} processes},
  year      = {2005},
  number    = {4},
  pages     = {411--424},
  volume    = {11},
  doi       = {10.1080/13873950500068567},
  publisher = {Informa UK Limited},
}

@Article{Byrd1995,
  author    = {Richard H. Byrd and Peihuang Lu and Jorge Nocedal and Ciyou Zhu},
  journal   = {SIAM Journal on Scientific Computing},
  title     = {A limited memory algorithm for bound constrained optimization},
  year      = {1995},
  number    = {5},
  pages     = {1190--1208},
  volume    = {16},
  doi       = {10.1137/0916069},
  publisher = {Society for Industrial \& Applied Mathematics (SIAM)},
}

@InProceedings{bonilla16,
  author    = {Edwin Bonilla and Daniel Steinberg and Alistair Reid},
  title     = {Extended and Unscented Kitchen Sinks},
  year      = {2016},
  address   = {New York, New York, USA},
  editor    = {Maria Florina Balcan and Kilian Q. Weinberger},
  pages     = {1651--1659},
  publisher = {PMLR},
  series    = {Proceedings of Machine Learning Research},
  volume    = {48},
  url       = {http://proceedings.mlr.press/v48/bonilla16.html},
}
\end{document}